\pdfoutput=1

\documentclass[preprint]{myelsarticle}

\usepackage{CJKutf8}

\usepackage{amssymb}

\usepackage[pdftex]{graphicx}

\newcommand{\deleted}[1]{}
\newcommand{\modified}[2]{#2}

\begin{document}

\begin{frontmatter}



\title{A Framework for Building Closed-Domain Chat Dialogue Systems}


\author[label1]{Mikio Nakano\footnote{Currently with C4A Research Institute, Inc.\\
E-mail address: mikio.nakano@c4a.jp}
}
\author[label2]{Kazunori Komatani}
\address[label1]{Honda Research Institute Japan Co., Ltd.\\
8-1 Honcho, Wako, Saitama 351-0188, Japan }
\address[label2]{The Institute of Scientific and Industrial Research\\
Osaka University\\
8-1 Mihogaoka, Ibaraki, Osaka 567-0047, Japan}

\begin{abstract}
This paper presents HRIChat, a framework for developing closed-domain
chat dialogue systems. Being able to engage in chat dialogues has been
found effective for improving communication between humans and
dialogue systems.  This paper focuses on closed-domain systems because
they would be useful when combined with task-oriented dialogue systems
in the same domain. HRIChat enables domain-dependent language
understanding so that it can deal well with domain-specific
utterances. In addition, HRIChat makes it possible to integrate state
transition network-based dialogue management and reaction-based
dialogue management. FoodChatbot, which is an application in the food
and restaurant domain, has been developed and evaluated through a user
study. Its results suggest that reasonably good systems can be
developed with HRIChat. This paper also reports lessons learned from
the development and evaluation of FoodChatbot.
\end{abstract}




\end{frontmatter}


\section{Introduction}

Dialogue systems are classified into task-oriented dialogue systems
and chat (or non-task-oriented) dialogue systems. Usually they are
studied differently, but combining them has been proposed
\cite{traum:iva05,nakano:humanoids06,lee:slt06} and has been found effective
in improving user impressions and the relationships with users
\cite{bickmore:tochi05,kobori:sigdial16,lucas:hri18}.

Most previously built chat dialogue systems are expected to
engage in open-domain dialogues. Recently they have been
studied intensively, and several competitions have been held
\cite{khatri:alexa18,dinan:arxiv19,higashinaka:iwsds19}.
However, when considering combining with a closed-domain
task-oriented dialogue system, chat dialogue systems in the
same domain are desired. 

One of the most important differences between open-domain chat
dialogue systems and closed-domain systems is that, while developing
one system may be enough for the former, developing a system for each
target domain is necessary for the latter. This means that we need to
make it easier to develop a system in the target domain. A framework
for developing closed-domain chat dialogue systems, therefore, is
desired.

As such a framework, this paper presents HRIChat,\footnote{\modified{*}{HRIChat was
previously called PyChat.}} which is implemented in Python. There are
two ideas behind HRIChat. First, it enables domain-specific language
understanding. This makes it possible for the system to extract
domain-specific user intention and information from user utterances
and it would lead to better responses.

The other idea is to combine state transition network-based dialogue management and
reaction-based dialogue management.  State transition network-based dialogue management
exploits a network for dialogues consisting of a small number of
turns.  Context is properly dealt with within the
dialogues. Typically, it is suitable for dialogues starting with a
system question.  On the contrary, reaction-based dialogue management
generates responses based on the preceding user utterance, without
taking into account longer context. Combining these types of
dialogue management modules enables the system to react to a variety of user
utterances and engage in dialogues in a context-dependent way.  This
is achieved by employing a multi-expert model \cite{nakano:kbs11} as
explained in Section~\ref{sec:overview}.

\modified{1-1}{Using HRIChat, we have built FoodChatbot, an
  application in the food and restaurant domain. It employs a graph
  database containing food and restaurant information. We conducted a
  user study with FoodChatbot, and its results show that FoodChatbot
  performs reasonably well and that HRIChat makes it possible to
  develop applications at such a level despite its simplicity. Note
  that we have not quantitatively compared HRIChat with another
  framework. This is because there is no existing framework for
  building closed-domain chatbots that we can compare with
  HRIChat. Instead we prove the concept of HRIChat through the
  development of FoodChatbot and the user study using it.}

This paper is organized as follows. Section~\ref{related} mentions
previous work related to closed-domain
chatbots. Section~\ref{proposed} describes HRIChat in detail and how to
build applications using HRIChat. Then Section~\ref{sec:app} explains
FoodChatbot, and Section~\ref{sec:evaluation} presents and discusses
the results of the user study. Section~\ref{sec:lessons} provides lessons
learned from the development and evaluation of FoodChatbot. Finally,
Section~\ref{conclusion} concludes the paper by mentioning future
work.

\section{Related Work}
\label{related}

One possible approach to building closed-domain chat dialogue systems
is to follow an approach to building open-domain systems.

There are a variety of open-domain chat dialogue systems. There are
rule-based systems that use user utterance patterns
\cite{weizenbaum1966eliza,wallace:book08}, retrieval-based systems
using examples
\cite{ritter:emnlp11,banchs:acl12demo,lowe:sigdial15,inaba:sigdial16},
and neural network-based dialogue generation models
\cite{vinyals:icmlws15,li:emnlp17,zhao:acl18}.  Higashinaka et
al. \cite{higashinaka:coling14} proposed a more complicated system
which uses a variety of response generation modules and selects one of
the outputs from those modules.
The systems that won the Alexa
Prizes\footnote{https://developer.amazon.com/alexaprize}
\cite{fang:alexa17,chen:alexa18} also exploit multiple
knowledge sources.

These approaches can be used also for building closed-domain systems,
but there are advantages and disadvantages in those approaches and one
disadvantage common to these approaches is that the use of context is
limited. We think handling context is important in closed-domain
systems, because the topic variation is limited so it is expected to
maintain the dialogue topic.

For task-oriented dialogue systems, state transition network-based (or finite-state
automaton-based) and frame-based dialogue management strategies are
often used \cite{jokinen:book09} and they can deal well with context,
so we decided to employ state transition network-based systems to deal with context
also for chat dialogue systems.

There have been a couple of studies on closed-domain chat dialogue
systems.  Sugiyama et al. \cite{sugiyama:slud18:e} built a system that
performs state transition network-based dialogue management and stores dialogue
contents that were shared with the user. It won first prize
at the dialogue system live competition held in Nov., 2018
\cite{higashinaka:iwsds19}. The system built by Bernsen et
al. \cite{bernsen:avi04} also employs several networks for dialogue
management.  HRIChat employs the same type of state transition network-based dialogue
management. Storing and accessing dialogue contents are also
possible. HRIChat supports not only state transition network-based dialogue management
but also simpler response generation based on reaction-based dialogue
management.

\begin{figure}
\centering
\includegraphics[width=0.7\columnwidth]{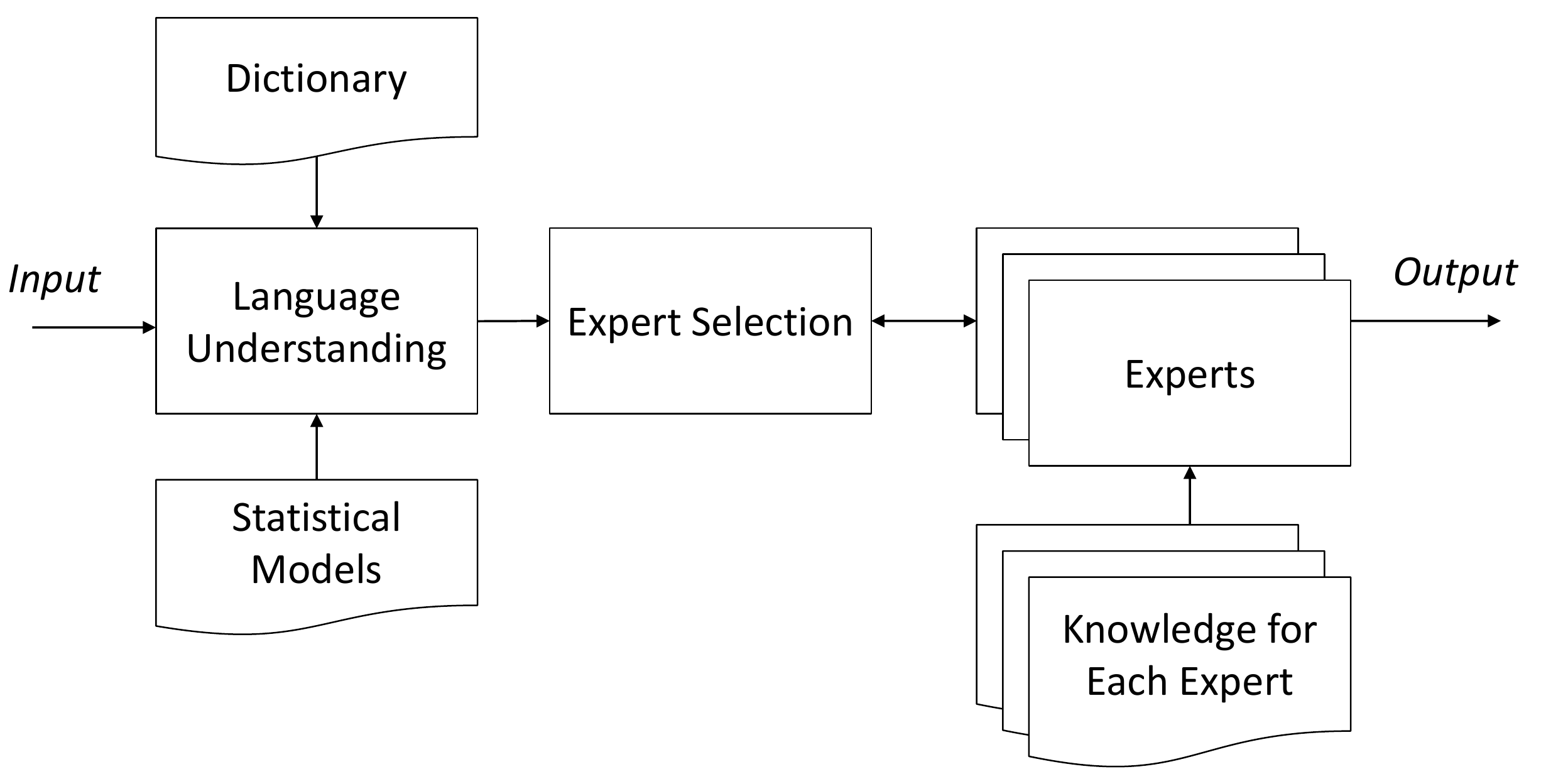}
\caption{Module architecture.}
\label{fig:arch}
\end{figure}

\section{HRIChat: the Proposed Framework}
\label{proposed}

\subsection{Multi-Expert Model}
\label{sec:overview}

HRIChat is based on a multi-expert model
\cite{nakano:kbs11}.  It features multiple experts each of
which manages dialogues in a different way using different
dialogue knowledge.  When the user inputs an utterance, the
language understanding module generates its semantic
representation.  Then it is sent to all the experts, and
each expert returns a score which indicates how likely it
should deal with the user input. The expert that returned
the highest score is activated, updates its internal state
based on the semantic representation, and generates a system
action. The system action can include an ``expert
activation'' command. In that case, the specified expert is
newly activated and generates another system action, and it
is executed after the execution of the system action
generated by the original expert.

Figure~\ref{fig:arch} depicts the module architecture. For
implementing experts, two expert classes, {\bf response expert} class
and {\bf small-talk network expert} (or network expert in short) class
are prepared. There can be only one instance of the response expert
class, and there can be multiple network experts. The response expert
performs reaction-based dialogue management using various types of
knowledge.  A network expert engages in dialogues with a small number
of turns, based on a state transition network.  Details of these
experts will be described later.

One of the advantages of the multi-expert model is that it is
possible to incorporate experts with any dialogue strategies
such as frame-based task-oriented dialogue management,
although we have used only the above two expert classes in
our application FoodChatbot described in Section~\ref{sec:app}.

\begin{figure}{\footnotesize
\begin{itemize}
\item[(1)] Determine the set of dialogue act types and the set of slot
  classes for language understanding.
\item[(2)] Prepare a set of example user utterances for training the
  statistical models for language understanding.
\item[(3)] Build a dictionary and implement functions to access it.
\item[(4)] Implement functions used in dialogue knowledge. 
\item[(5)] Write dialogue knowledge for the response expert and network experts. 
\item[(6)] Implement functions for each expert to select system actions.
\item[(7)] Write a configuration file that specifies files and
  parameters including the following:
\begin{itemize} 
\item Dialogue knowledge files
\item Files including developer-implemented functions
\item Parameters for statistical language understanding
\item Parameters for expert selection
\end{itemize}
\item[(8)] Implement functions called after selecting actions and understanding 
    user input (we call them {\em hooks}. This task is optional because these functions are mainly used for 
subtle dialogue control).
\end{itemize}
}
\caption{Tasks for application development.}

\label{fig:tasks}
\end{figure}

\subsection{Tasks for Application Development}

Figure~\ref{fig:tasks} lists the tasks required for developing an
application using HRIChat. These tasks can be done without much
expertise in natural language processing and dialogue systems, since
each task is simple unless the developers try to implement complicated
dialogue strategies. 

Below we explain these tasks in a rough explanation of
how an application works.

\subsection{Processes in Applications}

\subsubsection{Language Understanding}

The language understanding module assigns one of the
predefined dialogue act types to input user utterance and
extracts slots.  Differently from ordinary language
understanding, we use two kinds of dialog act types. One is a
coarse-grained type and the other is fine-grained. We call
the former {\em supertype} and the latter just {\em
  type}. 

Typically there are about 20 supertypes. Examples of supertypes are
{\em greet}, {\em acknowledge}, {\em ask-yes-no-question}, and {\em
  request-information}.  Types are domain-dependent and there can be
hundreds of types.  The set of supertypes, types, and slot classes
need to be defined by the developers (Figure~\ref{fig:tasks} (1)).  How
many slots in each kind appear in each user utterance is determined
depending on the type. The kinds of slots are domain dependent.

For example, the following utterance:
\begin{quote}
(a) Did you have sushi yesterday?
\end{quote}
is converted into the following semantic representation:
\begin{quote}
\begin{verbatim}
supertype: "ask-yes-no-question"
type:      "ask-if-system-ate"
slots:     [time-event="yesterday", food-drink="sushi"]
\end{verbatim}
\end{quote}

Each user utterance is first split into words using a morphological
analyzer.\footnote{When there are more than one sentence in the user
  input, the current version understands only the last sentence,
  although we plan to change this to understanding whole the user
  input.}  Then slot extraction and type/supertype prediction are
performed using statistical models. If the score of a type or
supertype prediction result is below a threshold, ``UNKNOWN'' is
assigned.  The statistical models are trained from a set of example
utterances prepared by the developers (Figure~\ref{fig:tasks} (2)).
For each example utterance, a type and a supertype are assigned and
slots are marked.

We assume that there is a dictionary which contains entries for each
class of slots. Each entry has spelling variations, synonyms, and
alternative terms (we call them {\em alternative names}).  If an
extracted slot value is one of the alternative names, it is replaced by the
entry in the semantic representation. HRIChat incorporates
developer-defined dictionary access functions.

Before the slot extraction using a statistical model, to better
extract words in the dictionary, the language understanding module
tries to match the input utterance with a short pattern having
dictionary entries.  If the matching succeeds, its result is used as
the slot extraction result.  To enable this, the developers need to
implement a dictionary and functions to access it
(Figure~\ref{fig:tasks} (3)).

Although each type belongs to one of the supertypes, type and
supertypes are independently predicted and their consistencies are not
considered, because they are not used at the same time. On the
contrary, we set a restriction on the relation between the type and
the kinds of slots. Among the 5-best type prediction results whose
scores are higher than the threshold, a type that is consistent with
the extracted slots is selected. We set this restriction because slot
values are used in generating responses using response pairs which
will be described in Section~\ref{sec:response-expert}.

After language understanding, extracted slot values are automatically
stored in predefined variables.  For example, after understanding
utterance (a) above, variables {\tt time-event1} and {\tt food-drink1}
are respectively set to \verb+"yesterday"+ and \verb+"sushi"+. Here
the suffix ``{\tt 1}'' means that it is the value of the leftmost food-drink
slot. This is necessary because there can be multiple slots in the
same class in one utterance as ``tea'' and ``coffee'' in ``Which do
you like, tea or coffee?". Those values are cleared after the system
makes the subsequent utterance.

It is possible to use other variables for
storing contextual information and using it.
For example, variable
{\tt topic} can be used by setting the value of other variables as follows:
\begin{quote}
\begin{verbatim}
topic = food-drink1
\end{verbatim}
\end{quote}
Alternatively, a symbol can be set to a variable directly.
\begin{quote}
\begin{verbatim}
topic = "noodle"
\end{verbatim}
\end{quote}
The values of these non-slot variables are not cleared unless
explicitly cleared. These are used for handling long contexts.

\subsubsection{System Action Realization}

Before explaining the process of experts and dialogue
knowledge for them, we explain the process of system action
realization which is common to experts.  Dialogue knowledge
in each expert has a different form depending on the class
of the expert, but it has system action descriptions, and
the system action that the expert outputs is realized from
one of the descriptions.
Each system action description consists of zero or one
condition, zero or more system utterances, zero or more
variable setting statements, and zero or one expert
activation statement. Below is an example.

\begin{quote}
\begin{verbatim}
label: "like-pizza"
condition: isPizza(food1)
utterance: "pizza is good"
utterance: "I like Hawaiian pizza"
setting-variable: topic="hawaiian pizza"
expert-activation: 
      expert-id="network1", 
      initial-state="ask-favorite-pizza"
\end{verbatim}
\end{quote}

The label is used in action selection when the developers
want to prioritize some actions explicitly. 

A condition consists of a developer-defined Boolean function
(Figure~\ref{fig:tasks} (4)) and arguments.  Arguments are
symbols or variables.  When there is a condition and it is
not satisfied, this system action is not realized.

When there are two or more system utterances, their
concatenation is presented to the user. 

A system utterance can include variables and function calls.  For
example, the following utterance description (b) is realized as system
utterance (c) when the value of {\tt *food1*} is {\tt "BBQ chicken
  pizza"} and {\tt get\_similar\_food("BBQ chicken pizza")} returns
{\tt "smoked chicken pizza"}.
\begin{itemize}
\item[(b)] {\tt "*food1* is great. Do you also like *get\_similar\_food(food1)*?"}
\item[(c)] {\tt "BBQ chicken pizza is great. Do you also like smoked chicken pizza?"}
\end{itemize}
Functions embedded in utterances need to be implemented by the
developers (Figure~\ref{fig:tasks} (4)).
If one of the variables is not set or one of the functions
is not defined, the utterance is not realized, and then the
system action is not realized.

\begin{figure}[t]
\begin{quote}
\noindent
{\small (a)}\\
\includegraphics[width=0.7\columnwidth]{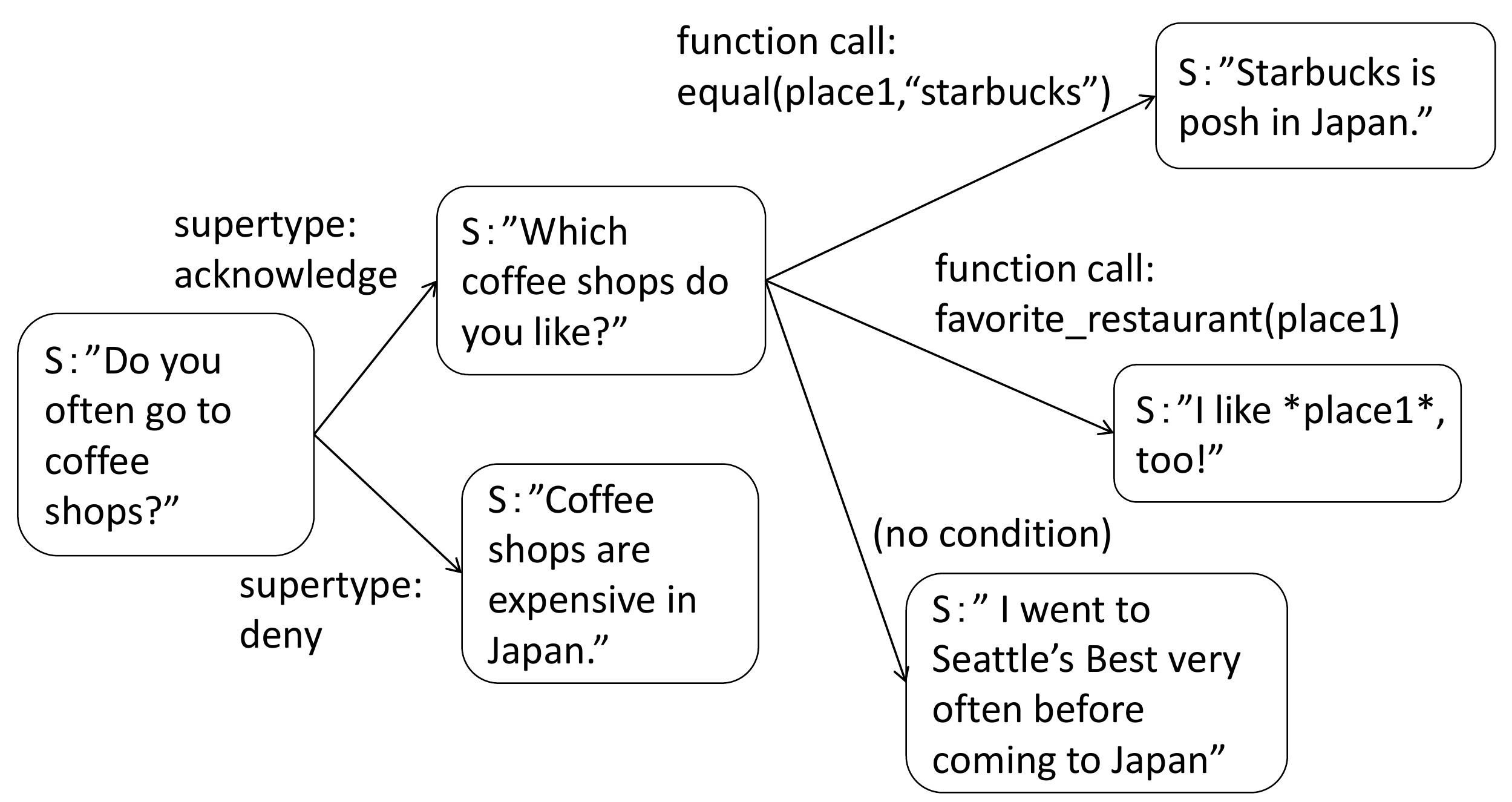}

\noindent
{\small
\begin{tabbing}
(b) \= System: \=Do you often go to coffee shops?\\
    \> User: \>Yes.\\
    \> System: \>Which coffee shops do you like?\\
    \> User: \>I like Tully's.\\
    \> System: \>I like Tully's, too! 
\end{tabbing}
}

\end{quote}
\caption{Sample network for the small-talk network expert and a
  dialogue example that can be generated from it.}
\label{fig:network}
\end{figure}

When there is an expert activation statement, the specified
expert is activated after utterance generation and variable
settings are finished.  Arguments can be passed to
the expert to be activated.  Typically, an action in the
response expert activates a network expert.  At this time,
the initial state of the network expert is specified as an
argument. The activated expert generates a system
action. Therefore two actions from the original expert and
the activated expert are performed sequentially.

\subsubsection{Small-Talk Network Expert}

A network expert utilizes a developer-defined network as the dialogue
knowledge for dialogue management (Figure~\ref{fig:tasks} (5)). It
consists of states and transitions.  Each state has one or more system
action descriptions and zero or more transitions.  Each transition has
conditions on the input user utterance and a destination
state. Conditions are the supertype of the user utterance, type of the
user utterance, or a function call (the function must be a
Boolean). Figure~\ref{fig:network} (a) illustrates an example
network. Rounded rectangles represent states and arrows
transitions. From this network, dialogue (b) can be generated.

When a network expert is activated, it realizes one of the
system action descriptions for the current state and outputs
its result. Then, when the subsequent user input comes, one
of the transitions whose conditions are satisfied
is selected and its destination state becomes the system's
new state.  Transitions are ordered, that is, the 
conditions are checked in the order of transitions and the
first one whose condition is satisfied is selected. There
can be transitions without any conditions. Such a transition is
selected regardless of the user input.

When there is no transition whose condition is satisfied,
the expert is deactivated and another expert is responsible
to select the system action. Even if there is a transition
whose conditions are satisfied, the expert might be
deactivated by the expert selector.

\begin{figure}
{\small
\begin{itemize}
\item[(a)] Example of response pairs:
\begin{verbatim}
 user-utterance-type: "ask-if-system-likes-food"
 system-action:
   condition: like(food-drink1)
   utterance: "Yes, I like *food-drink1* very much!"
\end{verbatim}

\item[(b)] A dialogue generated from (a) if {\tt like("sukiyaki")} returns {\tt "true"}
\begin{quote}
    User: Do you like {\em sukiyaki}?\\
    System: Yes, I like {\em sukiyaki} very much!
\end{quote}

\item[(c)] Example of default responses:
\begin{verbatim}
 user-utterance-supertype: "request-information"
 system-action: utterance: "Well, I have no idea."
\end{verbatim}
\item[(d)] A dialogue generated from (c): 
\begin{quote}
    User: Tell me more about that.\\
    System: Well, I have no idea.
\end{quote}
\end{itemize}
}
\caption{Knowledge examples for the response expert.}
\label{fig:response}
\end{figure}

\subsubsection{Response Expert}
\label{sec:response-expert}

The response expert exploits the following five types of
knowledge written by the developers (Figure~\ref{fig:tasks}
(5)).

A {\bf response pair} consists of a user utterance
type and a set of system action descriptions. This is for
responding based on precise language understanding results of
the user utterance. For example, Figure~\ref{fig:response}
(a) yields (b).

A {\bf default response} consists of a user
utterance supertype and a set of system action
descriptions. This makes it possible to respond based on the
rough classification results of the user utterance, even if
precise language understanding is not possible. For example,
Figure~\ref{fig:response} (c) yields (d).

An {\bf example response} consists of a user utterance example and a
system action description. This allows retrieval-based response
generation \cite{ritter:emnlp11}, which finds the user utterance
example that is the most similar to the inputted user utterance and
returns its corresponding system action. We employed tf-idf to
calculate the similarity for its simplicity. If the similarity is
lower than a set threshold, then no system action description is
listed.

A {\bf related response} consists of a topic word and one
system action description. When one of the extracted slot
values matches its topic word, its system action description
is listed.

\begin{figure}
{\small
\begin{itemize}

\item {\bf if} a network expert is activated and it is the first
  expert selection since the activation, \\ 
  {\bf then} select the network expert.

\item {\bf else if} response obligation exists, that is, the
  predicted supertype of the user utterance is one of the
  specific supertypes (e.g., {\em ask-yes-no-question} and
  {\em request-information}) and its score is above a threshold, \\
  {\bf then} select the response expert.

\item {\bf else if} a network expert is activated, and all the
  conditions of one of the transitions are satisfied, \\ {\bf then}
  select the network expert.

\item {\bf else if} there is at least one system action
  candidate realized from a response pair or an example
  response in the response expert, \\
  {\bf then} select the response expert.

\item {\bf else if} a network expert is activated and there is a
  transition having no conditions, \\ 
{\bf then} select the network expert.

\item {\bf else} select the response expert.

\end{itemize}
}
\caption{Expert selection algorithm.}
\label{fig:expert-selection}
\end{figure}

A {\bf non-response} consists of only one system action description.
This allows the system to respond or activate a network expert even if
language understanding fails.

Using these types of knowledge, the expert lists system
action descriptions, and realizes them to obtain system
action candidates. Then one action is selected using a
developer-implemented action selection function
(Figure~\ref{fig:tasks} (6)).

\subsubsection{Expert Selection}
\label{sec:expert_selection}

As mentioned earlier, the expert is selected based on the
scores that the experts return.  The default scores were
determined so that a network expert is deactivated depending
on how likely the response expert is suitable for dealing
with the user input and how likely the network expert should
continue the dialogue.  The default scores were determined
by trial and error in building the example
application. Figure~\ref{fig:expert-selection}
shows the algorithm corresponding to the default scores.

\subsection{Design Guidelines}

Together with the specification of HRIChat, we wrote
guidelines for knowledge
descriptions. Figure~\ref{fig:guideline} shows ones worth
noting.

\begin{figure}\small
\begin{itemize}
\item Make the system have consistency in its knowledge,
  utterance content, and linguistic style.
 \item Avoid too many response pairs so that the responses based on 
           misunderstanding do not occur.
\item Avoid too long dialogues based on one small-talk network so that
  the system does not stick to one small topic.
\item Try to avoid system utterances that may induce user questions so
  that the system does not fail to respond to unexpected user questions.
\item Prepare utterances in network experts to
  respond naturally even if the system cannot understand the user's answer to a system
  question.
\end{itemize}
\caption{A part of the design guidelines.}
\label{fig:guideline}
\end{figure}

\subsection{Implementation of HRIChat}

HRIChat is implemented in Python. Dialogue knowledge needs to be
written in XML but we also prepared a tool to generate XML files from
knowledge written in Microsoft Excel files.  Currently HRIChat supports
only Japanese, although we are planning to port it to other languages.
In addition, the current version of HRIChat deals with only text input
and output.

We employed MeCab \cite{kudo:emnlp04} for morphological analysis, and
used NEologd\footnote{https://github.com/neologd/mecab-ipadic-neologd}
for the dictionary for MeCab for the application described in
Section~\ref{sec:app}.  Slot extraction is based on sequential
labeling using IOB tagging and Conditional Random Fields
(CRF). HRIChat uses CRFsuite \cite{okazaki:crfsuite07} for the
implementation, and it uses commonly used features such as unigram and
bigram of the surface form, original form, and part of speech of the
word.  Supertype and type prediction is based on logistic regression
of scikit-learn \cite{pedregosa:jmlr11} using bag-of-words features,
which are original forms of words and question marks. \modified{1-3}{
  Although more advanced techniques such as deep neural network-based
  methods
  \cite{mesnil:taslp15,hakkani-tur:16,chen:arxiv19,gupta:sigdial19}
  can be used for achieving better language understanding
  performances, we employed simpler methods since the main purpose of
  this paper is to propose a new framework for building chat dialogue systems and
  we wanted to avoid computationally intensive methods to make HRIChat
  applications work in various computational environments. }

\section{FoodChatbot: An Example Application in the Food and Restaurant Domain}
\label{sec:app}

We developed {\em FoodChatbot}, an application in the food and
restaurant domain. \modified{*}{Note that FoodChatbot is not developed for
evaluating HRIChat. 
The performance of FoodChatbot shows how well an
application built with HRIChat can chat with naive users and what
problems remain. }

\subsection{System Character Design}

It is not possible to make the system answer to a variety of
questions concerning foods and restaurants by preparing a
comprehensive knowledge base.  So we designed the system
character so that it becomes natural that the system does
not know some food and restaurants and often ask questions.
The character is {\em Sophia}, a female American who
recently came to Japan and is interested in foods in Japan.

\begin{table}
{\small
\begin{center}
\begin{tabular}{|l|p{5cm}|p{5cm}|}
\hline {\bf Class}& {\bf Instances} & {\bf Example}
\\ \hline food-drink & food and drink names,
ingredients, cuisine types, meal type & sushi, coffee,
potato, Chinese, breakfast \\ place & place names,
restaurants, shops & home, New York, McDonald's
\\ time-event & time, event & morning, summer, Thanksgiving
\\ \hline
\end{tabular}
\end{center}
}
\caption{Slot classes.}
\label{tab:entities}
\end{table}

\subsection{Knowledge Base}
FoodChatbot uses a knowledge graph as a backend database. It
also works as the dictionary for language understanding. We
used three slot classes, namely, {\em food-drink}, {\em
  place}, and {\em time-event} (Table~\ref{tab:entities}). 
Instances in these classes
are represented as entities in the knowledge
graph. We limited the number of classes so that the
accuracy of the sequence labeling-based slot extraction
becomes high enough.

The database also includes the relationship between these
entities. For example, there are relations that {\em panna
  cotta}'s cuisine type is {\em Italian} and that {\em poke}
is a specialty of {\em Hawaii}. In addition, properties of
foods and drinks such as tastes and temperatures are
represented as relations.

There are entities whose labels are ``person'' for representing system
characters and users. Their knowledge, liking and experiences are also
represented as relations, although user information is not extracted
by the current FoodChatbot.  For example, that ``Sophia likes ramen''
is represented as a relation.

The initial database contains 10,291 entities (food-drink:
4,172, place: 5,186, time-event: 901, taste: 24,
temperature: 7, person: 1) and 27,899 relations (knowledge on
foods and drinks: 15,852, the system's personal information:
12,047).  We used Neo4j Community Edition\footnote{{\tt
    https://neo4j.com/}} for the database management. We
implemented functions for accessing the database that are
used in the dialogue knowledge.

\subsection{Session Topic}
In each session, the system determines one topic and chats on the
topic mainly by asking questions concerning it. We call it {\em
  session topic}.  An initial system utterance is prepared for each
session topic. When the system starts the session, it selects the
session topic and chooses the initial system utterance accordingly.

\begin{table}
{\small
\begin{center}
\begin{tabular}{|l|r|}
\hline
{\bf Type of knowledge} & {\bf Number}  \\ \hline
response pair & 98 \\
default response & 16 \\
example response & 3,124\\
related response & 451 \\ 
non-response & 164 \\ \hline
\end{tabular}
\end{center}
}
\caption{Amount of knowledge for the response expert.}
\label{tab:response-knowledge-amount}
\end{table}

\subsection{Dialogue Knowledge}
Table~\ref{tab:response-knowledge-amount} shows the amount
of knowledge for the response expert.  There is one network
expert, and it has 3,025 states and 2,938 transitions.

\subsection{Dialogue Strategy}
The basic dialogue strategy of FoodChatbot is that it
mainly engages in network dialogues and that sometimes the
response expert generates responses.  The network expert has
a number of sub-networks each of which starts from a state
asking a question to the user.  A list of such sub-networks
is assigned to each session topic, and they are used in a
predefined order.  The network expert is activated only from
the non-responses in the response expert. This strategy is
implemented in the dialogue knowledge, the action selector
for the response expert, and the hook functions ({\em c.f.},
Figure~\ref{fig:tasks} (8)).  Expert selection is done based
on the default scores.

\subsection{Language Understanding}
There are 16 supertypes and 332 types. There are three slot
classes as mentioned earlier. CRF for slot extraction was
trained with 21,983 utterances. These are user
utterances in the logs of dialogues between recruited users
and an older version of FoodChatbot.  Slots were manually
annotated for each utterance.  Logistic regression functions
for type and supertype prediction were trained with 9,904
utterances. They were generated from 1,603 developer-written
example utterances by replacing their
slots with similar words.\footnote{Whether two entities are
  similar or not is defined on the graph database.} We did
not use the utterances in the dialogue logs because we found
type and supertype annotation is not easy. We used different
thresholds and parameters for type prediction and
supertype prediction to avoid errors in type prediction, but
we do not explain this since it is too detailed.

\section{Evaluation of FoodChatbot}
\label{sec:evaluation}

\begin{figure}
\begin{center}
\includegraphics[width=0.6\columnwidth]{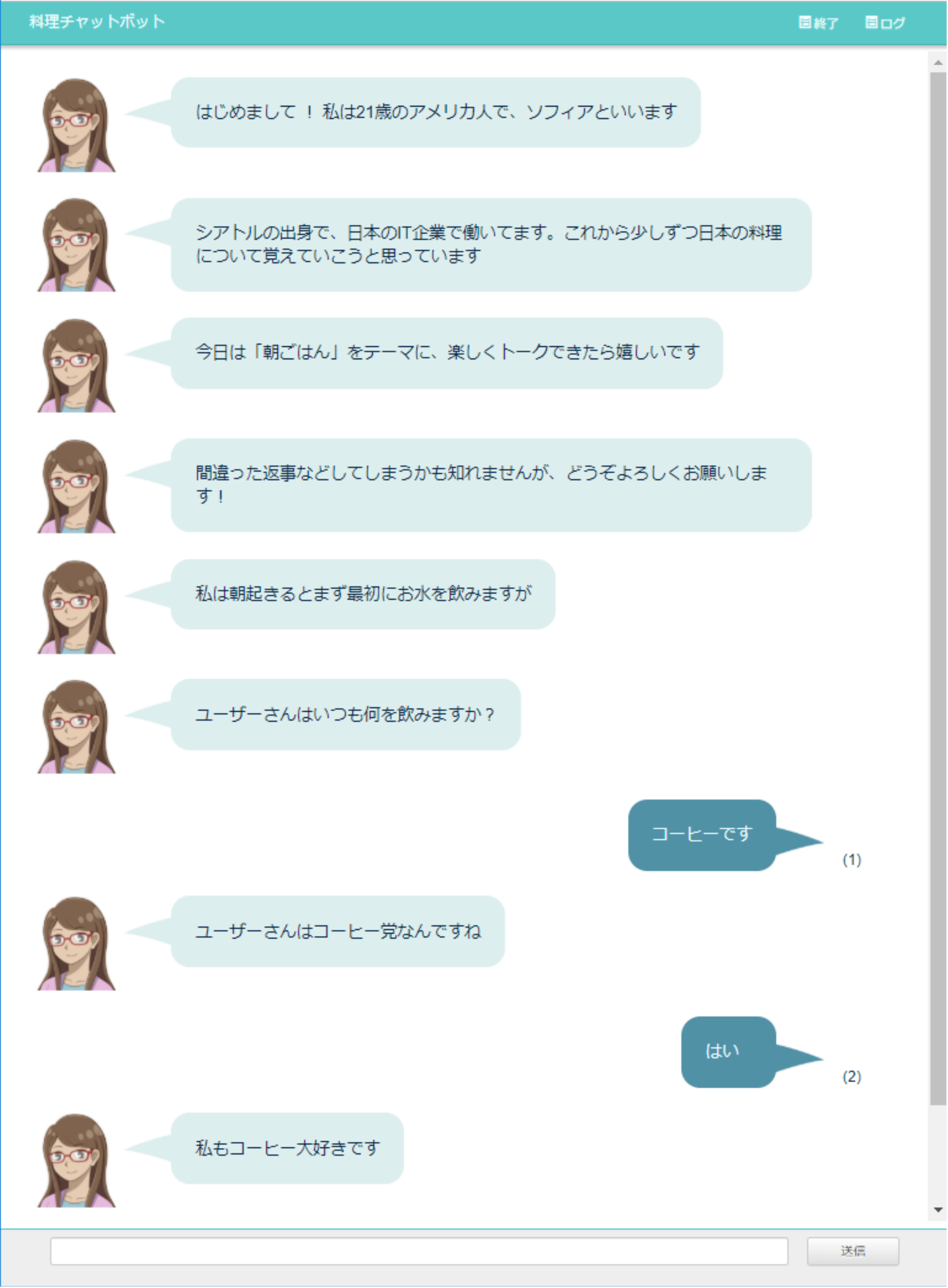}
\end{center}

{\small
Translations: \\
\begin{tabular}{lp{11cm}}
System: & Hi, I'm Sophia, I'm a 21 year-old American. \\
System: & I'm from Seattle, and I work for an IT company in Japan. I'm thinking about learning foods in Japan little by little.\\
System: & Today, let's have fun talking about ``breakfast''.\\
System: & Forgive me if I say something strange.\\
System: & I drink water when I get up.\\
System: & What do you drink?\\
User: & I drink coffee.\\
System: & You are a coffee-drinker.\\
User: & Yes.\\
System: & I like coffee very much, too.
\end{tabular}
}
\caption{A snapshot of the Web chat interface.}

\label{fig:snapshot}
\end{figure}

\modified{1-2}{We conducted a user study with FoodChatbot which is an
  application of our HRIChat framework in the food and restaurant
  domain. The purpose of this user study is to investigate how well an
  application of our framework interacts with real users. In other
  words, this is a kind of proof of concept of HRIChat.
}

\subsection{Procedure of the User Study}

\modified{*}{We recruited 62 people via a crowd-sourcing company CrowdWorks
Inc.}\footnote{https://crowdworks.jp/} The participants accessed the
dialogue server from a Web browser on their own devices such as PCs
and smartphones. Figure~\ref{fig:snapshot} shows a snapshot.

The participants were requested to engage in eight sessions
within eight days, and not to engage in more than two
sessions a day.

Each session started with a system utterance and consisted of 15
system turns and 15 user turns. The last user turn was not understood
and the system said that it terminated the session. After finishing
each session, the participants were asked to evaluate the dialogue by
rating eight items on 5-point Likert-scales (1: strongly disagree, 2:
disagree, 3: neutral, 4: agree, 5: strongly agree).

We prepared 10 session topics. The session topic for the
first session for each user was ``breakfast''.  In the
remaining sessions, the session topic was chosen randomly,
but the participants did not chat on the same session topic
twice.

\begin{table}{\small
\begin{center}
\begin{tabular}{|l|r|} \hline
{\bf Expert (knowledge type)} & {\bf Frequency} \\ \hline
initial $\rightarrow$ network & 1.00 \\
response (response pair) & 0.68 \\
response (default response) & 1.08 \\
response (example response) & 0.37 \\
response (related response) & 0.47 \\ 
response (non-response) $\rightarrow$ network & 2.29 \\
network & 9.11 \\ \hline
total & 15.00 \\ \hline
\end{tabular}
\end{center}
}
\caption{Average frequency of responses made by each expert
  with each type of knowledge in each session. A right arrow
  denotes the activation of an expert, and ``initial'' means
  the system's first utterance. }
\label{tab:expert}
\end{table}

\begin{figure}
{\footnotesize
\begin{tabular}{rp{4.5cm}p{1.5cm}rp{4cm}}
\hline\hline
& English translation & Expert and knowledge type & Score 
&Original Japanese utterance \\ \hline\hline
S1 &
  Hi, user! It's good to have {\em nabe} (Japanese hot pot) in winter!
$\mid$ I'm wondering if you can tell me about {\em nabe} dishes Japan prides itself on.
$\mid$ Recently, I often cook {\em kimuchi nabe} (Korean pickle hot pot).
$\mid$ What kind of {\em nabe} do you cook often?&
initial $\rightarrow$ network & &
\begin{CJK}{UTF8}{ipxm}ハーイ、ユーザーさん！ 冬はなんといってもお鍋ですよね！\end{CJK}
$\mid$ \begin{CJK}{UTF8}{ipxm}今回は日本が誇る「鍋料理」について、いろいろ教えてもらえると嬉しいです\end{CJK}
$\mid$ \begin{CJK}{UTF8}{ipxm}私はここ最近キムチ鍋をよくします\end{CJK}
$\mid$ \begin{CJK}{UTF8}{ipxm}ユーザーさんがよくするお鍋は何ですか？\end{CJK}	\\ \hline 
U2 &
I often cook chicken salt {\em nabe}.& 
& &
\begin{CJK}{UTF8}{ipxm}鶏肉の塩なべをよく作ります\end{CJK} \\ \hline 
S3 &
It will make you warm. $\mid$ What foodstuff do you put in it?  &
network&  2 & 
\begin{CJK}{UTF8}{ipxm}あったまりそう\end{CJK} $\mid$ 
\begin{CJK}{UTF8}{ipxm}具材はどんなのが入るのでしょうか\end{CJK} \\ \hline
U4 & 
I put Chicken thigh meat, Chinese cabbage, mushrooms, and bean sprouts. &
& &
\begin{CJK}{UTF8}{ipxm}鶏もも肉、白菜、キノコ類、もやしを入れます\end{CJK} \\ \hline 
S5 &
You can have a lot of vegetables. &
network  &  2 &
\begin{CJK}{UTF8}{ipxm}野菜もたっぷりとれますね\end{CJK}  \\ \hline
U6 &
Putting ginger will warm you more. &
& &
\begin{CJK}{UTF8}{ipxm}生姜を入れるとさらに体が温まります\end{CJK} \\ \hline 
S7 &
Do you have {\em zousui} (rice simmered in the soup) to finish up with the pot?&
network &  2 &
\begin{CJK}{UTF8}{ipxm}お鍋の〆は雑炊ですか？\end{CJK} \\ \hline
U8 &
I have {\em zousui} more often. &
& &
\begin{CJK}{UTF8}{ipxm}雑炊の方がおおいです。\end{CJK}   \\ \hline 
S9 & 
I often have {\em zousui} to finish up with {\em nabe}.  &
response (related-response) & -1 &
\begin{CJK}{UTF8}{ipxm}お鍋のシメは雑炊が多いです\end{CJK} \\ \hline
U10 & 
{\em Udon} (wheat noodle) is also good. &
& &
\begin{CJK}{UTF8}{ipxm}うどんもおいしいですよね\end{CJK}  \\ \hline 
S11  & 
{\em Udon}. I understand. &
response (response-pair) & 2 &
\begin{CJK}{UTF8}{ipxm}うどんですよね、分かります\end{CJK}\\ \hline
U12 & 
{\em Nabe} with Chinese cabbage and pork is also tasty. &
& &
\begin{CJK}{UTF8}{ipxm}白菜と豚肉の鍋もおいしいです\end{CJK}  \\ \hline 
S13 & 
I heard some people have {\em nabe} every day in winter.
$\mid$ How many times did you have {\em nabe} this winter?&
response (non-response) $\rightarrow$ network   &  1 &
\begin{CJK}{UTF8}{ipxm}冬は毎日お鍋って人もいるようですね\end{CJK}
$\mid$ \begin{CJK}{UTF8}{ipxm}ユーザーさんは、今シーズン何回くらいお鍋食べましたか？\end{CJK} \\ \hline 
U14 & 
I think I had {\em nabe} around 50 times because I had it almost every day. &
& &
\begin{CJK}{UTF8}{ipxm}ほぼ毎日食べてたので50回くらいだとおもいます\end{CJK}  \\ \hline 
S15 &
I have {\em nabe} almost once a week this month. &
network &  1 &
\begin{CJK}{UTF8}{ipxm}私は今月は毎週のように食べてます\end{CJK} \\ \hline
U16 &
Do you change the taste every day? &
& &
\begin{CJK}{UTF8}{ipxm}毎日味は変えているんですか？\end{CJK}    \\ \hline 
S17 &
Having {\em nabe} warms both the room and body. &
network &   -2 &
\begin{CJK}{UTF8}{ipxm}お鍋をすると、お部屋も身体もポカポカになりますね\end{CJK} \\ \hline
U18 &
Cooking pots is good because it's easy.&
& &
\begin{CJK}{UTF8}{ipxm}お鍋は調理も簡単でいいですよね\end{CJK} \\ \hline 
\end{tabular}
}
\caption{An example dialogue. The symbol $\mid$ is a delimiter for
  creating multiple balloons on the chat interface. S and U in the
  left most column respectively mean system and users utterances.
  Score is the evaluation score explained in
  Section~\ref{sec:results}.}
\label{fig:example1}
\end{figure}

\begin{figure}[t]
{\footnotesize
\begin{tabular}{rp{4.5cm}p{1.5cm}rp{4cm}}
\hline\hline
& English translation & Expert and knowledge type & Score 
&Original Japanese utterance \\ \hline\hline
S19 &
I usually have {\em nabe} alone $\mid$ but it's also fun to have it with many people. &
network &  2 &
\begin{CJK}{UTF8}{ipxm}いつもは一人鍋ですが\end{CJK}$\mid$
\begin{CJK}{UTF8}{ipxm}大勢で取り合って食べるのも楽しいですね\end{CJK}   \\ \hline
U20 & 
I want to have {\em chanko nabe} (a kind of hot pot famous as a Sumo wrestlers' meal) 
with many people.& &
&
\begin{CJK}{UTF8}{ipxm}大勢でちゃんこ鍋とか食べたいです\end{CJK}    \\ \hline 
S21 &
Let me join next time! &
network   & 2 &
\begin{CJK}{UTF8}{ipxm}今度誘ってくださいね！\end{CJK} \\ \hline
U22 & 
Definitely! &
& &
\begin{CJK}{UTF8}{ipxm}是非！\end{CJK} \\ \hline 
S23 &
By the way, sliced meat for {\em shabushabu} (a type of hot pot) is very thin.
$\mid$ I was surprised that the meat is very thin when I had {\em shabushabu} for the first time.
&
response (non-response) $\rightarrow$ network   &  2 &
\begin{CJK}{UTF8}{ipxm}ところで、しゃぶしゃぶのお肉って薄いですね\end{CJK}
$\mid$ \begin{CJK}{UTF8}{ipxm}初めてしゃぶしゃぶ食べた時、あまりの薄さに驚きました\end{CJK} \\ \hline
U24 & 
We can't slice pork that thinly by ourselves. &
& &
\begin{CJK}{UTF8}{ipxm}自分では切れない薄さですよね\end{CJK}     \\ \hline 
S25 &
I haven't seen such thinly sliced meat in the States.&
network   & 2 &
\begin{CJK}{UTF8}{ipxm}アメリカであんなに薄いお肉、見たことなかったです\end{CJK}  \\ \hline
U27 &
It's tasty though it's thin.&
& &
\begin{CJK}{UTF8}{ipxm}薄いのにおいしいんですよね\end{CJK}     \\ \hline
S28 &
Probably we can't cook {\em shabushabu} if meat is thickly sliced. &
 network   & 2 &
\begin{CJK}{UTF8}{ipxm}お肉が厚かったら、しゃぶしゃぶってできなそうですね\end{CJK} \\ \hline
U29 &
 It's rare boiled meat. &
 & &
\begin{CJK}{UTF8}{ipxm}レアなゆで肉ですものね\end{CJK}     \\ \hline 
S30 & 
Sauces for {\em shabushabu} is also good. &
network  & 1 &
\begin{CJK}{UTF8}{ipxm}しゃぶしゃぶのタレも美味しいですよね\end{CJK}  \\ \hline
\end{tabular}
}
\caption{An example dialogue (continued).}
\label{fig:example2}
\end{figure}

\subsection{Collected Dialogues}

We analyzed the results of the 53 participants (30 females and 23
males whose ages ranged from 20s to 50s) who seriously engaged in eight sessions
and made 15 utterances for all of the sessions.  The average number of
words in a system turn was 17.4\footnote{This includes initial system
  utterances which tend to be long.} and that in a user turn was 7.3.
Table~\ref{tab:expert} shows the average frequencies of the selected
experts and knowledge types used to generate responses for all 424
sessions. Most of the system utterances were made by the network
expert and sometimes the response expert made utterances.

\modified{1-1}{The dialogue server worked in real time. Based on the
  server logs, we found that the average response time (the elapsed time
  between a user input and the subsequent system output) was 124.7ms
  over all dialogues for the 53 participants and the standard
  deviation was 89.6ms. The maximum response time was 844ms. 
}

Figures~\ref{fig:example1} and \ref{fig:example2} show an example of
the collected dialogues. This dialogue's average score of system
utterance evaluation results described below is close to the median
among all dialogues. In S1, the system tells the user the session
topic of this dialogue is {\em nabe} (Japanese hot pot). Then the
network expert proceeds the dialogue by asking questions or telling
its opinions. There are two problematic system utterances.  In S9 the
system said the same thing as what the user said in U8.  This is
because the system could not understand that U8 was an acknowledgement
of the system question S7. U8 does not include a simple linguistic
expression corresponding to ``yes'', so it is not easy to recognize
its intention and this is among our future work.  In S17, the system
ignored the user's question in U16. This is because, although
top-scored supertype prediction results for U16 is
``request-information'', its score is below the threshold, so the
system did not detect response obligation and the network expert was
selected. Better scoring for supertype prediction results and
thresholding for response obligation detection are also among our
future work.

\subsection{Results}
\label{sec:results}

We evaluated language understanding using the data for 10 randomly
selected participants. \modified{2-2}{We used only a part of the data because 
annotating correct labels requires a lot of manpower and the data for the 10 participants
is large enough to roughly evaluate the language understanding performance.}
The accuracy of supertype prediction was 44.9\% (503/1120). For type
prediction, since it is not easy to annotate correct results, we
calculated its accuracy for the utterances whose type prediction
results are not ``UNKNOWN'', and it was 41.1\% (216/525). We guess
these poor performances were due to the difficulty in consistent
type/supertype annotation on example sentences.  The F1 score of slot
extraction was 74.2\%. It could be improved if we had a better
guideline for the annotations on the training data. Unlike
task-oriented dialogues, annotations are not easy for non-task
oriented dialogues. Note that language understanding failures and
errors are not always problematic because the experts do not select
actions based only on language understanding results.

Table~\ref{tab:impression} shows the results of the questionnaire. We
focus mainly on Item 1 which is used in the preliminary selection of
the dialogue system live competition \cite{higashinaka:iwsds19}.
FoodChatbot's results are between its third and fourth
systems.\footnote{Note that ``strongly agree'' is 5 in our evaluation
  while it is 1 in the competition.} However, since the evaluation
settings are different in several points,\footnote{First, in the
  preliminary selection of the dialogue system live competition, some
  participants chatted with multiple (5.8 on average) systems which
  participated in the competition \cite{higashinaka:slud18:e}, while
  our participants used only one system. Second, we used a Web
  interface, not Telegram. Third, in our evaluation, the 15th user
  utterance was not understood and responded in each session.}  it is
not appropriate to directly compare with systems which participated in
the competition. In addition, there might not be statistical
significances in differences.  Nevertheless, this suggests we can
build an application which performs reasonably well using HRIChat.

We calculated the linear regression function of each item
for each user.  For all items, the averages of their slopes
are positive (Table~\ref{tab:impression}). This means the
participants' impression did not get worse as they engaged
in a greater number of sessions.

Table~\ref{tab:impression} also shows the correlation
between Item 1 and the remaining items.  From this, we find
that fun and naturalness are crucial for the participants'
willingness to chat with the system again, and that the
system's ability to understand user utterances is also
important.

\begin{table*}\footnotesize
\begin{center}
\begin{tabular}{|r|p{3.5cm}|r|r|r|r|}\hline
\multicolumn{2}{|l|}{\bf Questionnaire item} & {\bf All (S.D.)} & {\bf First (S.D.)} & {\bf Slope (S.D.)} & \multicolumn{1}{|l|}{\bf Correlation} \\ 
\multicolumn{2}{|l|}{} &  &  &  &  \multicolumn{1}{|l|}{\bf with Item 1}  \\  \hline
1 & I'm willing to chat with the system again. & 3.56 (0.99) & 3.62 (0.95) & 0.01 (0.09) & - \\ 
2 & The dialogue was fun. & 3.48 (1.03) & 3.55 (1.01) & 0.01 (0.11) & 0.86 \\ 
3 & *The system was friendly. & 4.09 (0.88) & 3.79 (0.93) & 0.07 (0.11) & 0.39 \\ 
4 & The system understood my utterances. & 3.13 (1.07) & 3.21 (1.01) & 0.01 (0.16) & 0.62 \\ 
5 & The dialogue was natural. & 3.11 (1.05) & 3.02 (1.08) & 0.05 (0.14) & 0.66 \\ 
6 & *The dialogue went well. & 3.03 (1.11) & 2.83 (1.01) & 0.07 (0.16) & 0.45\\ 
7 & *The system was polite. & 4.31 (0.78) & 4.40 (0.74) & 0.02 (0.11) & 0.40 \\ 
8 & *The system did not often change the topic. & 3.50 (1.03) & 2.81 (1.08) & 0.11 (0.14) & 0.35\\ \hline
\end{tabular}\\
\end{center}
\noindent
{\bf All}: the average score over all sessions.\\
{\bf First}: the average score over the first sessions for each user.\\
{\bf Slope}: the average of the slopes of the linear regression function for each user.\\
\caption{User impression scores. For the items marked with ``*'', reversed questions were asked to the participants.
}
\label{tab:impression}
\end{table*}

We also evaluated the system utterances except the first turn in a
five-point scale (-2: very bad, -1: bad, 0: neutral, 1: good, 2: very
good). One evaluator rated all data and another evaluator rated data
for 10 participants. 
The agreement rate between two evaluators in
Krippendorff's $\alpha{}$ (interval scale) was 0.84, which is high
enough, so we used the scores of the first evaluator. We found the
average of the mean score of the system utterances for each session is
1.06 (S.D.: 0.64), so we think that the system utterances are
reasonably good. However, we also found that correlation between the
mean score of the system utterances and the questionnaire item 1 is
0.40, so there seem to be other factors that affect Item 1. This issue
needs further investigation.

\section{Lessons Learned}
\label{sec:lessons}

Through the development and evaluation of FoodChatbot, we
learned several lessons.

First, one reason for the poor performance of language
understanding is that the annotations of types and slots are
not consistent in the training data.  We need to find a way
to establish annotation guidelines for each application
domain and a better way to design the set of types and slot
classes.

We found the developers who were in charge of writing dialogue
knowledge want to fix the order of system questions to avoid
contradictions. So we implemented action selection functions so that
the system asks questions in a predefined order, but implementing such
functions is not simple. In addition, we found it is not easy to avoid
contradiction between system utterances by consistently building
dialogue knowledge and the database, so it is desired that
contradictions are automatically detected and system actions are
properly selected.

HRIChat allows the developers to use variables and
functions, making it possible to use a variety of contextual
information and knowledge in the database so that dialogues
become more natural and interesting. However, it does not
seem easy to use those features. We think it would be
useful to show examples in which those features are
effectively used.

\section{Concluding Remarks}
\label{conclusion}

This paper presented HRIChat, a framework for building closed-domain
chat dialogue systems. It makes it possible to employ domain-specific
language understanding, and also allows combining reaction-based
dialogue management and state transition network-based dialogue
management.  Through the evaluation results of FoodChatbot, an
application of HRIChat in the food and restaurant domain, it is found
that a system whose performance is reasonably good can be developed
with HRIChat. Although there is much room to improve, the current
status of HRIChat is worth reporting considering the evaluation results
of FoodChatbot and lessons learned from it development.

There are many existing technologies that are yet to be incorporated
into HRIChat. First, language understanding performance could be
improved by more advanced techniques such as deep neural network-based
methods
\cite{mesnil:taslp15,hakkani-tur:16,chen:arxiv19,gupta:sigdial19}.
\modified{2-1}{However, since such methods are computationally
  intensive, they might require enhancing the hardware, need longer
  time for model training, and make real-time responses difficult. So
  we need to investigate the trade-off between the performance
  improvement and the increase in the computational costs. It would be also effective to
  exploit contextual information in finding example responses in the
  response expert
  \cite{banchs:acl12demo,lowe:sigdial15,inaba:sigdial16}.}  \modified{*}{In addition,
  extracting the user's personal information
  \cite{tsunomori:iwsds19} and interests \cite{inaba:sigdial18} is worth considering}
  because such information is crucial for generating better
  utterances, e.g., avoiding asking the user again what he/she already
  said, and avoiding asking about what he/she is not interested in.

Expert selection and action selection within the experts
could be improved using the dialogue data collected with an
initial version of application. However, how to build
annotated training data without much effort and expertise is
yet to be explored.

We also plan to use HRIChat for building another
application to investigate how easy or difficult it is to
build a new system from scratch. Also we have a plan to
build a system that can engage in both chat and
task-oriented dialogues such as restaurant search. This can
be easily done by incorporating an expert for task-oriented
dialogues.

\section*{Acknowledgments}

We would like to thank all people who contributed to the development
of HRIChat and FoodChatbot.
This work is funded by Honda Research Institute Japan Co., Ltd.

\bibliography{../bib/sigdial-macro,../bib/list}

\begin{thebibliography}{10}
\expandafter\ifx\csname url\endcsname\relax
  \def\url#1{\texttt{#1}}\fi
\expandafter\ifx\csname urlprefix\endcsname\relax\def\urlprefix{URL }\fi
\expandafter\ifx\csname href\endcsname\relax
  \def\href#1#2{#2} \def\path#1{#1}\fi

\bibitem{traum:iva05}
D.~Traum, W.~Swartout, S.~Marsella, J.~Gratch, Fight, flight, or negotiate:
  Believable strategies for conversing under crisis, in: Proceedings of the
  International Conference on Intelligent Virtual Agents 2005 (IVA-2005), 2005,
  pp. 52--64.
\newline\urlprefix\url{https://link.springer.com/chapter/10.1007/11550617\_5}

\bibitem{nakano:humanoids06}
M.~Nakano, A.~Hoshino, J.~Takeuchi, Y.~Hasegawa, T.~Torii, K.~Nakadai, K.~Kato,
  H.~Tsujino, A robot that can engage in both task-oriented and
  non-task-oriented dialogues, in: Proceedings of the 6th IEEE-RAS
  International Conference on Humanoid Robots (Humanoids 2006), 2006, pp.
  404--411.
\newline\urlprefix\url{http://dx.doi.org/10.1109/ICHR.2006.321304}

\bibitem{lee:slt06}
C.~Lee, S.~Jung, M.~Jeong, G.~G. Lee, Chat and goal-oriented dialog together: A
  unified example-based architecture for multi-domain dialog management, in:
  Proceedings of the 2006 IEEE Spoken Language Technology Workshop (SLT-2006),
  2006.
\newline\urlprefix\url{http://dx.doi.org/10.1109/SLT.2006.326788}

\bibitem{bickmore:tochi05}
T.~W. Bickmore, R.~W. Picard,
  \href{https://doi.org/10.1145/1067860.1067867}{Establishing and maintaining
  long-term human-computer relationships}, ACM Transactions on Computer-Human
  Interaction 12~(2) (2005) 293--327.
\newblock \href {https://doi.org/10.1145/1067860.1067867}
  {\path{doi:10.1145/1067860.1067867}}.
\newline\urlprefix\url{https://doi.org/10.1145/1067860.1067867}

\bibitem{kobori:sigdial16}
T.~Kobori, M.~Nakano, T.~Nakamura,
  \href{https://www.aclweb.org/anthology/W16-3646}{Small talk improves user
  impressions of interview dialogue systems}, in: Proceedings of the 17th
  Annual Meeting of the Special Interest Group on Discourse and Dialogue
  (SIGDIAL 2016), 2016, pp. 370--380.
\newline\urlprefix\url{https://www.aclweb.org/anthology/W16-3646}

\bibitem{lucas:hri18}
G.~M. Lucas, J.~Boberg, D.~R. Traum, R.~Artstein, J.~Gratch, A.~Gainer,
  E.~Johnson, A.~Leuski, M.~Nakano,
  \href{https://doi.org/10.1145/3171221.3171258}{Getting to know each other:
  The role of social dialogue in recovery from errors in social robots}, in:
  Proceedings of the 2018 ACM/IEEE International Conference on Human-Robot
  Interaction (HRI-2018), 2018, pp. 344--351.
\newblock \href {https://doi.org/10.1145/3171221.3171258}
  {\path{doi:10.1145/3171221.3171258}}.
\newline\urlprefix\url{https://doi.org/10.1145/3171221.3171258}

\bibitem{khatri:alexa18}
C.~Khatri, B.~Hedayatnia, A.~Venkatesh, J.~Nunn, Y.~Pan, Q.~Liu, H.~Song,
  A.~Gottardi, S.~Kwatra, S.~Pancholi, M.~Cheng, Q.~Chen, L.~Stubel,
  K.~Gopalakrishnan, K.~Bland, R.~Gabriel, A.~Mandal, D.~Hakkani-Tur, G.~Hwang,
  N.~Michel, E.~King, R.~Prasad, Advancing the state of the art in open domain
  dialog systems through the {Alexa} {Prize}, in: Proceedings of the Alexa
  Prize 2018, 2018.

\bibitem{dinan:arxiv19}
E.~Dinan, V.~Logacheva, V.~Malykh, A.~H. Miller, K.~Shuster, J.~Urbanek,
  D.~Kiela, A.~Szlam, I.~Serban, R.~Lowe, S.~Prabhumoye, A.~W. Black, A.~I.
  Rudnicky, J.~Williams, J.~Pineau, M.~Burtsev, J.~Weston,
  \href{http://arxiv.org/abs/1902.00098}{The {Second} {Conversational}
  {Intelligence} {Challenge} {(ConvAI2)}}, arXiv preprint 1902.00098 (2019).
\newline\urlprefix\url{http://arxiv.org/abs/1902.00098}

\bibitem{higashinaka:iwsds19}
R.~Higashinaka, K.~Funakoshi, M.~Inaba, Y.~Tsunomori, T.~Takahashi, R.~Akama,
  Dialogue system live competition: identifying problems with dialogue systems
  through live event, in: Proceedings of the Tenth International Workshop on
  Spoken Dialogue Systems Technology (IWSDS-2019), 2019.

\bibitem{nakano:kbs11}
M.~Nakano, Y.~Hasegawa, K.~Funakoshi, J.~Takeuchi, T.~Torii, K.~Nakadai,
  N.~Kanda, K.~Komatani, H.~G. Okuno, H.~Tsujino,
  \href{https://doi.org/10.1016/j.knosys.2010.08.004}{A multi-expert model for
  dialogue and behavior control of conversational robots and agents},
  Knowledge-Based Systems 24~(2) (2011) 248--256.
\newblock \href {https://doi.org/10.1016/j.knosys.2010.08.004}
  {\path{doi:10.1016/j.knosys.2010.08.004}}.
\newline\urlprefix\url{https://doi.org/10.1016/j.knosys.2010.08.004}

\bibitem{weizenbaum1966eliza}
J.~Weizenbaum, \href{https://doi.org/10.1145/365153.365168}{{ELIZA}--a computer
  program for the study of natural language communication between man and
  machine}, Communications of the ACM 9~(1) (1966) 36--45.
\newblock \href {https://doi.org/10.1145/365153.365168}
  {\path{doi:10.1145/365153.365168}}.
\newline\urlprefix\url{https://doi.org/10.1145/365153.365168}

\bibitem{wallace:book08}
R.~S. Wallace, The anatomy of {A.L.I.C.E.}, in: R.~Epstein, G.~Roberts,
  G.~Beber (Eds.), Parsing the Turing Test: Philosophical and Methodological
  Issues in the Quest for the Thinking Computer, Springer, 2008, pp. 181--210.

\bibitem{ritter:emnlp11}
A.~Ritter, C.~Cherry, W.~B. Dolan,
  \href{http://www.aclweb.org/anthology/D11-1054}{Data-driven response
  generation in social media}, in: Proceedings of the 2011 Conference on
  Empirical Methods in Natural Language Processing (EMNLP-2011), 2011, pp.
  583--593.
\newline\urlprefix\url{http://www.aclweb.org/anthology/D11-1054}

\bibitem{banchs:acl12demo}
R.~E. Banchs, H.~Li, \href{http://www.aclweb.org/anthology/P12-3007}{{IRIS}: a
  chat-oriented dialogue system based on the vector space model}, in:
  Proceedings of the 50th Annual Meeting of the Association for Computational
  Linguistics (ACL-2012): System Demonstrations, 2012, pp. 37--42.
\newline\urlprefix\url{http://www.aclweb.org/anthology/P12-3007}

\bibitem{lowe:sigdial15}
R.~Lowe, N.~Pow, I.~V. Serban, J.~Pineau,
  \href{http://aclweb.org/anthology/W/W15/W15-4640.pdf}{The {Ubuntu} dialogue
  corpus: A large dataset for research in unstructured multi-turn dialogue
  systems}, in: Proceedings of the 16th Annual Meeting of the Special Interest
  Group on Discourse and Dialogue (SIGDIAL 2015), 2015, pp. 285--294.
\newline\urlprefix\url{http://aclweb.org/anthology/W/W15/W15-4640.pdf}

\bibitem{inaba:sigdial16}
M.~Inaba, K.~Takahashi,
  \href{http://aclweb.org/anthology/W/W16/W16-3648.pdf}{Neural utterance
  ranking model for conversational dialogue systems}, in: Proceedings of the
  17th Annual Meeting of the Special Interest Group on Discourse and Dialogue
  (SIGDIAL 2016), 2016, pp. 393--403.
\newline\urlprefix\url{http://aclweb.org/anthology/W/W16/W16-3648.pdf}

\bibitem{vinyals:icmlws15}
O.~Vinyals, Q.~Le, A neural conversational model, in: Proceedings of the Deep
  Learning Workshop at ICML 2015, 2015.

\bibitem{li:emnlp17}
J.~Li, W.~Monroe, T.~Shi, S.~Jean, A.~Ritter, D.~Jurafsky,
  \href{https://aclanthology.info/papers/D17-1230/d17-1230}{Adversarial
  learning for neural dialogue generation}, in: Proceedings of the 2017
  Conference on Empirical Methods in Natural Language Processing (EMNLP-2017),
  2017, pp. 2157--2169.
\newline\urlprefix\url{https://aclanthology.info/papers/D17-1230/d17-1230}

\bibitem{zhao:acl18}
T.~Zhao, K.~Lee, M.~Esk{\'{e}}nazi,
  \href{https://aclanthology.info/papers/P18-1101/p18-1101}{Unsupervised
  discrete sentence representation learning for interpretable neural dialog
  generation}, in: Proceedings of the 56th Annual Meeting of the Association
  for Computational Linguistics (ACL-2018), 2018.
\newline\urlprefix\url{https://aclanthology.info/papers/P18-1101/p18-1101}

\bibitem{higashinaka:coling14}
R.~Higashinaka, K.~Imamura, T.~Meguro, C.~Miyazaki, N.~Kobayashi, H.~Sugiyama,
  T.~Hirano, T.~Makino, Y.~Matsuo,
  \href{http://aclweb.org/anthology/C/C14/C14-1088.pdf}{Towards an open-domain
  conversational system fully based on natural language processing}, in:
  Proceedings of the 25th International Conference on Computational Linguistics
  (COLING-2014), 2014, pp. 928--939.
\newline\urlprefix\url{http://aclweb.org/anthology/C/C14/C14-1088.pdf}

\bibitem{fang:alexa17}
H.~Fang, H.~Cheng, E.~Clark, A.~Holtzman, M.~Sap, M.~Ostendorf, Y.~Choi, N.~A.
  Smith, {Sounding Board} -- {University of Washington's} {Alexa Prize}
  submission, in: Proceedings of the Alexa Prize 2017, 2017.

\bibitem{chen:alexa18}
C.-Y. Chen, D.~Yu, W.~Wen, Y.~M. Yang, J.~Zhang, M.~Zhou, K.~Jesse, A.~Chau,
  A.~Bhowmick, S.~Iyer, G.~S.~R. Cheng, A.~Bhandare, Z.~Yu, Gunrock: Building a
  human-like social bot by leveraging large scale real user data, in:
  Proceedings of the Alexa Prize 2018, 2018.

\bibitem{jokinen:book09}
K.~Jokinen, M.~McTear, Spoken Dialogue Systems, Morgan and Claypool Publishers,
  2009.

\bibitem{sugiyama:slud18:e}
H.~Sugiyama, H.~Narimatsu, M.~Mizukami, T.~Arimoto, Empirical study on
  domain-specic conversational dialogue system based on context-aware utterance
  understanding and generation, in: SIG-SLUD-B5-02, Japanese Society for
  Artificial Intelligence, 2018, pp. 118--123, (in Japanese).

\bibitem{bernsen:avi04}
N.~O. Bernsen, M.~Charfuelan, A.~Corradini, L.~Dybkj{\ae}r, T.~Hansen,
  S.~Kiilerich, M.~Kolodnytsky, D.~Kupkin, M.~Mehta,
  \href{https://doi.org/10.1145/989863.989951}{First prototype of
  conversational {H.C.} andersen}, in: Proceedings of the Working Conference on
  Advanced Visual Interfaces (AVI 2004), 2004, pp. 458--461.
\newblock \href {https://doi.org/10.1145/989863.989951}
  {\path{doi:10.1145/989863.989951}}.
\newline\urlprefix\url{https://doi.org/10.1145/989863.989951}

\bibitem{kudo:emnlp04}
T.~Kudo, K.~Yamamoto, Y.~Matsumoto,
  \href{http://www.aclweb.org/anthology/W04-3230}{Applying conditional random
  fields to {Japanese} morphological analysis}, in: Proceedings of the 2004
  Conference on Empirical Methods in Natural Language Processing (EMNLP-2004),
  2004, pp. 230--237.
\newline\urlprefix\url{http://www.aclweb.org/anthology/W04-3230}

\bibitem{okazaki:crfsuite07}
N.~Okazaki, {CRFsuite}: a fast implementation of conditional random fields
  {(CRFs)}, http://www.chokkan.org/software/crfsuite/ (2007).

\bibitem{pedregosa:jmlr11}
F.~Pedregosa, G.~Varoquaux, A.~Gramfort, V.~Michel, B.~Thirion, O.~Grisel,
  M.~Blondel, P.~Prettenhofer, R.~Weiss, V.~Dubourg, J.~VanderPlas, A.~Passos,
  D.~Cournapeau, M.~Brucher, M.~Perrot, E.~Duchesnay,
  \href{http://jmlr.org/papers/v20/17-100.html}{Scikit-learn: {Machine}
  learning in {Python}}, Journal of Machine Learning Research (2011)
  2825--2830.
\newline\urlprefix\url{http://jmlr.org/papers/v20/17-100.html}

\bibitem{mesnil:taslp15}
G.~Mesnil, Y.~Dauphin, K.~Yao, Y.~Bengio, L.~Deng, D.~Z. Hakkani-Tur, X.~He,
  L.~P. Heck, G.~Tur, D.~Yu, G.~Zweig,
  \href{https://doi.org/10.1109/TASLP.2014.2383614}{Using recurrent neural
  networks for slot filling in spoken language understanding}, IEEE/ACM
  Transactions on Audio, Speech and Language Processing 23~(3) (2015) 530--539.
\newblock \href {https://doi.org/10.1109/TASLP.2014.2383614}
  {\path{doi:10.1109/TASLP.2014.2383614}}.
\newline\urlprefix\url{https://doi.org/10.1109/TASLP.2014.2383614}

\bibitem{hakkani-tur:16}
D.~Hakkani{-}T{\"{u}}r, G.~T{\"{u}}r, A.~{\c{C}}elikyilmaz, Y.~Chen, J.~Gao,
  L.~Deng, Y.~Wang,
  \href{https://doi.org/10.21437/Interspeech.2016-402}{Multi-domain joint
  semantic frame parsing using bi-directional {RNN-LSTM}}, in: Proceedings of
  the 17th Annual Conference of the International Speech Communication
  Association (Interspeech 2016), 2016, pp. 715--719.
\newblock \href {https://doi.org/10.21437/Interspeech.2016-402}
  {\path{doi:10.21437/Interspeech.2016-402}}.
\newline\urlprefix\url{https://doi.org/10.21437/Interspeech.2016-402}

\bibitem{chen:arxiv19}
Q.~Chen, Z.~Zhuo, W.~Wang, \href{https://arxiv.org/abs/1902.10909}{{BERT} for
  joint intent classification and slot filling}, arXiv preprint
  arXiv:1902.10909 (2019).
\newline\urlprefix\url{https://arxiv.org/abs/1902.10909}

\bibitem{gupta:sigdial19}
A.~Gupta, J.~Hewitt, K.~Kirchhoff, Simple, fast, accurate intent classification
  and slot labeling for goal-oriented dialogue systems, in: Proceedings of the
  20th Annual Meeting of the Special Interest Group on Discourse and Dialogue
  (SIGDIAL 2019), 2019, pp. 46--55.
\newline\urlprefix\url{https://www.aclweb.org/anthology/W19-5906/}

\bibitem{higashinaka:slud18:e}
R.~Higashinaka, K.~Funakoshi, M.~Inaba, Y.~Tsunomori, T.~Takahashi, R.~Akama,
  The dialogue system live competition, in: SIG-SLUD-B5-02, Japanese Society
  for Artificial Intelligence, 2018, pp. 106--111, (in Japanese).

\bibitem{tsunomori:iwsds19}
Y.~Tsunomori, R.~Higashinaka, T.~Yoshimura, Y.~Isoda, Chat-oriented dialogue
  system that uses user information acquired through dialogue and its long-term
  evaluation, in: Proceedings of the Tenth International Workshop on Spoken
  Dialogue Systems Technology (IWSDS-2019), 2019.

\bibitem{inaba:sigdial18}
M.~Inaba, K.~Takahashi,
  \href{https://www.aclweb.org/anthology/W18-5004}{Estimating user interest
  from open-domain dialogue}, in: Proceedings of the 19th Annual Meeting of the
  Special Interest Group on Discourse and Dialogue (SIGDIAL 2018), 2018, pp.
  32--40.
\newline\urlprefix\url{https://www.aclweb.org/anthology/W18-5004}

\end{thebibliography}
\bibliographystyle{elsarticle-num}

\end{document}